\documentclass[11pt,a4paper]{llncs}

\usepackage{amssymb}    

\usepackage{amsmath}
\usepackage{tipa}
\usepackage{array}
\usepackage[caption=false,font=normalsize,labelfont=sf,textfont=sf]{subfig}
\usepackage{textcomp}
\usepackage{stfloats}
\usepackage{url}
\usepackage{graphicx}
\usepackage{fancyhdr}
\usepackage{geometry}
\newcommand{\keywords}[1]{\par\addvspace\baselineskip
\noindent\keywordname\enspace\ignorespaces#1}
\DeclareTextFontCommand\textphonetic{\ipafont}
\usepackage{hyperref}

\fancypagestyle{firstpage}{\fancyhf{}
}

\author{
Hadi Zaatiti\inst{1}
\and
Hatem Hajri\inst{2}
\and
Osama Abdullah\inst{1}
\and
Nader Masmoudi\inst{3,4}
}

\institute{%
Bio-Medical Imaging Core, Core Technology Platforms,\\
New York University Abu Dhabi, PO Box 129188, Abu Dhabi, UAE\\
\email{\{hadi.zaatiti, osama.abdullah\}@nyu.edu}
\and                         
 Institut de recherche technologique SystemX, 8 Avenue de la Vauve, 91120 Palaiseau, France.
\\
\email{hatem.hajri@irt-systemx.fr}
\and                         
NYUAD Research Institute, New York University Abu Dhabi,\\
PO Box 129188, Abu Dhabi, UAE
\and                         
Courant Institute of Mathematical Sciences,\\
New York University, 251 Mercer Street, New York, NY 10012, USA\\
\email{masmoudi@cims.nyu.edu}
}

\begin{document}

\title{\LARGE{Towards stable AI systems  for Evaluating Arabic Pronunciations}}

\maketitle

\thispagestyle{firstpage}

\begin{abstract}
Modern Arabic ASR systems such as wav2vec 2.0 excel at word- and sentence-level transcription, yet struggle to classify isolated letters. In this study, we show that this phoneme-level task, crucial for language learning, speech therapy, and phonetic research, is challenging because isolated letters lack co-articulatory cues, provide no lexical context, and last only a few hundred milliseconds. Recogniser systems must therefore rely solely on variable acoustic cues, a difficulty heightened by Arabic’s emphatic (pharyngealized) consonants and other sounds with no close analogues in many languages. This study introduces a diverse, diacritised corpus of isolated Arabic letters and demonstrate that state-of-the-art wav2vec 2.0 models achieve only 35 \% accuracy on it. Training a lightweight neural network on wav2vec embeddings raises performance to 65 \%. However, adding a small amplitude perturbation ($\epsilon$ = 0.05) cuts accuracy to 32 \%. To restore robustness, we apply adversarial training, limiting the noisy-speech drop to 9 \% while preserving clean-speech accuracy. We detail the corpus, training pipeline, and evaluation protocol, and release, on-demand, data and code for reproducibility. Finally, we outline future work extending these methods to word- and sentence-level frameworks, where precise letter pronunciation remains critical.
\keywords{Arabic letters pronunciation, adversarial training, classification}
\end{abstract}




\section{Introduction}

Accurately recognizing the pronunciation of isolated Arabic letters is a challenging task in speech recognition, compounded by the rich phonetic characteristics of Arabic and the variability introduced by different speakers. Arabic features a diverse set of phonemes, including several emphatic (pharyngealized) consonants and other sounds with no direct equivalent in many languages. These subtle phonetic distinctions can be difficult for automatic systems to capture, especially when presented with a single spoken letter devoid of any surrounding context. As a learning task, for non-native speakers, the challenge is even greater: many Arabic sounds (e.g., the pharyngeal \/'ayn\/ or the emphatic \/ṣ\/) are unfamiliar and often conflated with more common sounds, leading to mispronunciations. An isolated letter offers no lexical or syntactic context to aid recognition, so the system must rely entirely on acoustic cues that may vary widely across speakers and recording conditions. This lack of context, combined with inter-speaker variability and L2 pronunciation errors, makes high-accuracy classification of spoken Arabic letters a non-trivial problem. Indeed, even though Arabic is one of the most widely spoken languages globally, it remains relatively under-resourced in speech technology research, and fine-grained tasks like phoneme-level recognition pose significant hurdles \cite{Almekhlafi2022}. State-of-the-art speech recognition models such as \texttt{wav2vec~2.0} \cite{baevski2020wav2vec} have demonstrated impressive performance on word-level and sentence-level speech tasks by leveraging self-supervised learning on large corpora. However, these models are typically optimized for continuous speech and may not be directly adequate for isolated letter/phoneme classification. \texttt{wav2vec~2.0}’s powerful transformer-based architecture learns contextualized representations from long audio sequences; as a result, it expects and exploits surrounding context to aid recognition. 

In this paper, we are concerned with the problem of automatically classifying and scoring phoneme-level tasks, where each input is an extremely short utterance (often a single consonant or vowel sound), the lack of context can lead to suboptimal feature representations and confusion between acoustically similar letters. Recent studies have noted that applying \texttt{wav2vec~2.0} to phonetic classification requires careful adaptation for example, providing additional padding or context around a target phoneme to achieve reasonable accuracy \cite{Kim2024}. This indicates that a vanilla \texttt{wav2vec~2.0} model, if used without modification, might struggle to distinguish certain Arabic phonemes in isolation. In essence, models trained and tuned on high-level speech units do not inherently capture the fine-grained articulatory nuances needed for reliable letter-level recognition. This gap calls for dedicated approaches that bridge the difference in granularity, ensuring that each Arabic sound can be recognized on its own, even when spoken by learners with imperfect pronunciation. 

\textbf{Motivation.} Beyond the technical difficulty, the ability to correctly classify and score spoken Arabic letters has broad significance in several domains. In language learning and Computer-Assisted Pronunciation Training (CAPT), providing feedback on individual phoneme pronunciation is crucial for helping students master a second language. Arabic, in particular, is learned by millions of non-native speakers worldwide for both communicative and religious reasons \cite{Alrashoudi2025}. Precise automatic evaluation of a learner’s pronunciation at the phoneme level can enable personalized feedback detecting which specific Arabic letters a learner struggles with (e.g., distinguishing \textsubdot{h}\=a' from h\=a') and guiding them on how to improve. This is especially important because Arabic is a phonetic language where pronunciation errors at the letter level can propagate to word-level misunderstandings. Similarly, in both news broadcasting and clinical speech therapy contexts, training often begins with isolated sounds or letters whether to help reporters refine articulation or to support children with speech sound disorders before progressing to full words and sentences. Another critical domain is Quranic recitation and Tajweed, where correct pronunciation of each letter is mandated. Tajweed rules require delivering every letter of the Quran with its due articulation characteristics; even slight deviations in pronouncing a letter (for instance, not properly emitting the deep q\=af or the throaty \textsubdot{h}\=a') are considered errors that can alter meanings or reduce the recitation’s correctness.

\textbf{Related work.} There has been some prior work aimed at Arabic phoneme and letter classification, but it remains relatively limited, and several gaps persist. Early approaches to Arabic phoneme recognition often relied on traditional machine learning and acoustic features, achieving only modest success on restricted phoneme sets. More recently, researchers have started to apply deep learning to this problem. For example, Almekhlafi et al. (2022) introduced a dedicated Arabic Alphabet Phonetics Dataset (AAPD) and trained various deep neural network models for isolated letter classification \cite{Almekhlafi2022}. Their benchmark systems, using features like Mel-frequency cepstral coefficients (MFCCs) and architectures ranging from simple DNNs to CNNs (including a VGG-based model), reported high classification accuracy (on the order of 95\% for clean, native-speaker audio). This work demonstrates the feasibility of Arabic letter classification with deep learning, and provides a strong baseline. However, there are notable limitations. Many existing studies, including AAPD, focus primarily on data from native speakers or well-articulated audio; they may not fully capture the variability and accented pronunciations introduced by non-native learners. Moreover, these models are typically evaluated in ideal conditions, and their reliability can degrade in more realistic settings with noise, reverberation, or speaker accent shifts. 

Crucially, previous works have generally not addressed the robustness of Arabic phoneme classifiers i.e., how stable the predictions are when the input speech is slightly perturbed or when facing deliberate attempts to confuse the model. There is also a lack of exploration into integrating modern self-supervised speech models (like \texttt{wav2vec}) specifically for letter-level recognition prior successes in Arabic ASR mostly pertain to word or sentence recognition, or to detecting pronunciation errors within spoken phrases \cite{Alrashoudi2025}. This leaves a gap in understanding how well advanced ASR networks perform on minimal speech units, and whether additional techniques are needed to make them effective in that regime. It has been shown that perturbations in an audio signal can cause even state-of-the-art ASR systems to produce incorrect outputs or transcriptions \cite{Carlini2018}. Robustness to perturbations is crucial; models deployed in real-world scenarios, must handle background noise, microphone variation, and other distortions without confusion. 
A common method to guarantee robustness against multiple types of noise, is adversarial training. Recent surveys and studies highlight the vulnerabilities of current ASR models and exploring defenses to harden them against perturbations \cite{Huynh2022}. These techniques have  been well investigated to improve resilience of ASR models, but remain under-explored in analyzing the robustness of phoneme-level classifiers.

\textbf{Outline.} In this paper, we first construct a new dataset of spoken Arabic letters, collected from a diverse pool of speakers including both native Arabic speakers and non-native learners. We acquired a carefully curated corpus of isolated letter recordings. The data was collected, via a specially designed mobile and web app, from geographically diverse, gender- and age-balanced speakers; performed expert-level annotation to establish a high-fidelity ground truth. This dataset encompasses the diacritized Arabic consonant phonemes pronounced in isolation, providing a broad coverage of accents and pronunciation qualities.

Using this resource, we evaluate the performances of the state-of-the-art ASR model (\texttt{wav2vec~2.0})
on this dataset in the context of Arabic letter-level classification. Our analysis, showed poor results of the state-of-the-art model. We then fine-tune a neural network on this dataset and show that it highly outperformed the (\texttt{wav2vec~2.0})model. Furthermore, a core component of our work is a thorough stability analysis of the resulting models. We subject the classifiers to a range of perturbations to probe their robustness. This includes infusing the dataset with perturbations of various amplitudes, the latter modelling synthetic noise, minor time/frequency distortions, and generating adversarial examples that attempt to confuse the model into misclassification. By evaluating model performance under these conditions, we can identify specific vulnerabilities (e.g., particular letters that are easily mistaken for others under noise) and gauge improvements from any defense techniques we incorporate (such as adversarial training or data augmentation). The introduction of adversarial testing in the context of Arabic phoneme classification is, to our knowledge, a novel aspect of this work, shining light on an often-neglected dimension of model performance.



\section{Data collection}
We begin by describing the collection and filtering of isolated Arabic letter recordings, followed by the processing steps used to prepare the data for model training.

\subsection{Data acquisition}

Obtaining high-quality data for Arabic speech research presents multiple challenges. First, geographic dispersion of native speakers makes it difficult to gather consistent recordings, especially when aiming for balanced representation across regions, age groups, and genders. Second, annotation is particularly challenging due to the absence of reliable phonetic aligners for Arabic and the limitations of existing transcription models. Accurate labeling requires reference to a consistent pronunciation standard, which is not readily available. To overcome these barriers, we developed a web-based and a mobile based application to facilitate data acquisition and storage.
\begin{itemize}
    \item Web application:  \href{https://arabic-pronunciation.abudhabi.nyu.edu/}{arabic-pronunciation.abudhabi.nyu.edu}
    \item Mobile application: \href{https://play.google.com/store/apps/details?id=com.arabic_pronunciation.pronunciation_app}{Arabic Pronunciation Android app on Google Play}
\end{itemize}

\begin{figure}
    \centering
    \includegraphics[scale=0.5]{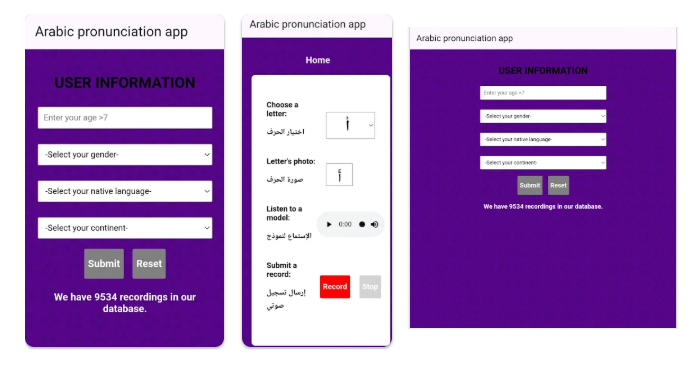}
    \caption{An android mobile app and a web app are developped to facilitate data acquisition of the ``Horouf" dataset, a model of each pronunciation is proposed for the user to follow before recording and submitting their pronunciation.}
    \label{fig:app}
\end{figure}

Figure \ref{fig:app} illustrates the graphical user interface of the mobile application which is similar to the web application. The user chooses a pronunciation model that is a syllabus comprising an Arabic consonant followed by a vowel, making each datapoint a diacritized letter.
Each recording is labelled as $l = (A,D) $ such that  $A$ is one of the Arabic letters and $D$ is one of the four diactrics: "fatha" /a/, "kasra" /i/, "damma" /u/, and "sokoon" (absence of any vowel). The total number of classes is therefore 112. The user records his pronunciation, reviews his recording, and then chooses to either submit or discard his contribution.  Additionally, meta-data related to the speaker (gender, age, nativeness, continent) are also collected. In its current state, the ``Horouf" (``Letters" in Arabic) dataset has a total number of approximately ten thousand recordings. The data is anonymous and safely stored on NYU Abu Dhabi secure servers. Arabic speakers are welcome to contribute to the dataset. 

\subsection{Pre-processing steps}

In this section we detail the pre-processing steps of the collected audio samples. Even though the user chooses the class before submitting his recording, the data is carefully annotated by linguistic experts and checked against the class to ensure phonetic accuracy and consistency, this step is the most lengthy and costly one. 
We perform the following processing steps:
\begin{itemize}
    \item Remove beginning and end silences. Silence detection is done by weighing the average signal energy measurement against a small threshold, followed by a manual verification.
    \item Manual labeling and cleaning of the data by Arabic language experts, any inaccurate pronunciation or non-conforming example to the provided model is discarded. On average, 11\% of the data was rejected either due to high noise levels or improper pronunciation. 
    \item Prior to being used for training, the data is split 80\% as training data and 20\% as test data.
    \item The 80 \%  (around 8K samples) were augmented using the following techniques: Gaussian white noise, Random pitch shift, Random time-stretch and circular time shift. The final number of samples after augmentation reaches 30K samples.
\end{itemize}

\subsection{Embedding of the raw audio files} Pre-training the raw waveform with \texttt{wav2vec~2.0} provides a powerful sequence of contextualized feature vectors. In our pipeline, we feed each 16\,kHz signal into the publicly released \texttt{wav2vec2-large}
\texttt{-XLSR-53-Arabic}  \cite{xlsr53_model} checkpoint, which is built on a 1024-dimensional Transformer encoder trained in a self-supervised fashion on 53 languages. The model returns a matrix $\mathbf{H} \in \mathbb{R}^{T \times 1024}$ whose rows correspond to 20\,ms speech frames, with $T$ the duration of the utterance and $\mathbf{h}_t$ the embedding vector at time $t$. We convert this variable-length matrix into a fixed-length utterance embedding vector $\mathbf{e}$ by taking the arithmetic mean over the time axis:

\[
\mathbf{e} = \frac{1}{T}\sum_{t=1}^{T} \mathbf{h}_t,
\]

yielding a single 1024-dimensional vector per file. Temporal mean-pooling is a simple yet effective aggregation strategy used in various speech tasks to produce utterance-level representations~\cite{zhang2017deep,snyder2018x}, preserving phonetic content while discarding timing variability and noise. Despite the incredibly short duration of the collected audio data from ``Horouf", the resulting embeddings inherit the multilingual phonetic knowledge encoded by \texttt{wav2vec}
\texttt{2.0} \cite{baevski2020wav2vec,conneau2020unsupervised}, making them well suited for downstream classification of isolated Arabic letters. The computed vectors are now used as input for all sequel operations described in the next sections.


\section{Transcription and data classification}

First we shed light on the identified gap in state of the art. The state of the art transcription model on the Arabic language (named "wav2vec2-large-xlsr-53-arabic") had a 37\% accuracy on our dataset to recognize the diacritized letters. As the transcription model could indicate multiple correct transcriptions of the same pronunciation, we performed the scoring of the model manually by Arabic language experts. The experts were instructed to be very generous while scoring the model: if the consonant part of the pronunciation is correct followed by the correct diacritic (``damma", ``fatha" or ``kasra", regardless whether or not a ``madd" (elongation) exists, the model transcription is graded as correct, and it is graded incorrect in all other situations.

\begin{figure}[ht]
    \centering
    \includegraphics[scale=0.4]{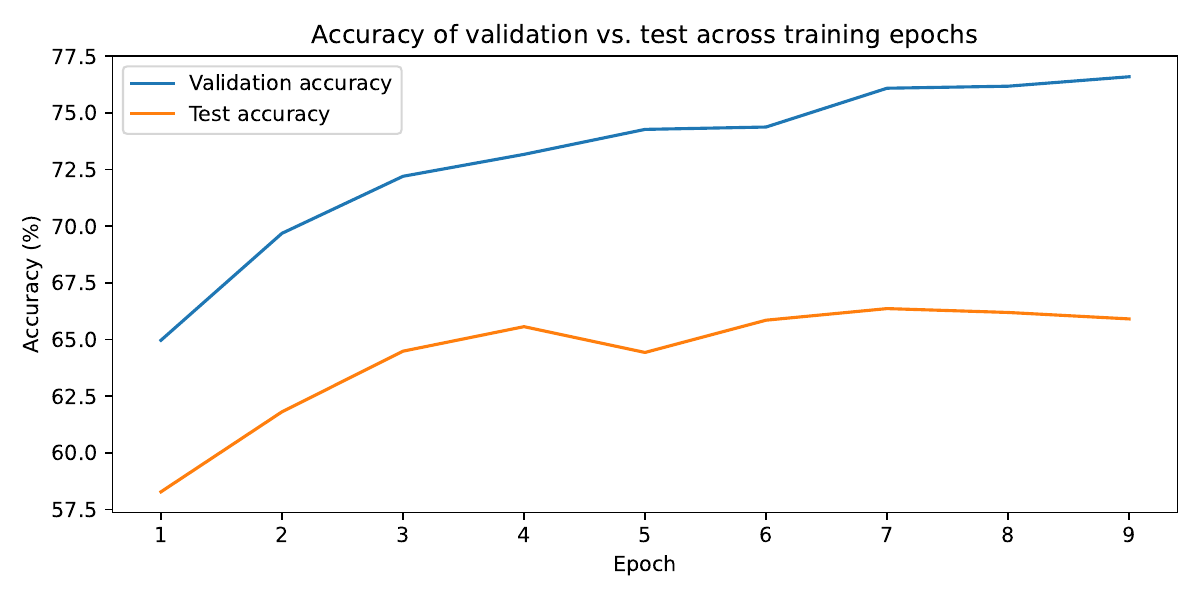}
    \caption{Accuracy of validation vs. test data from ``Horouf" of a classical multilayer perceptron (MLP) neural network over 9 epochs. State-of-the-art model showed 37\% accuracy on the test data.}
    \label{fig:mlp_classifier}
\end{figure}

\begin{figure*}[h]
    \centering
    \includegraphics[width=\textwidth]{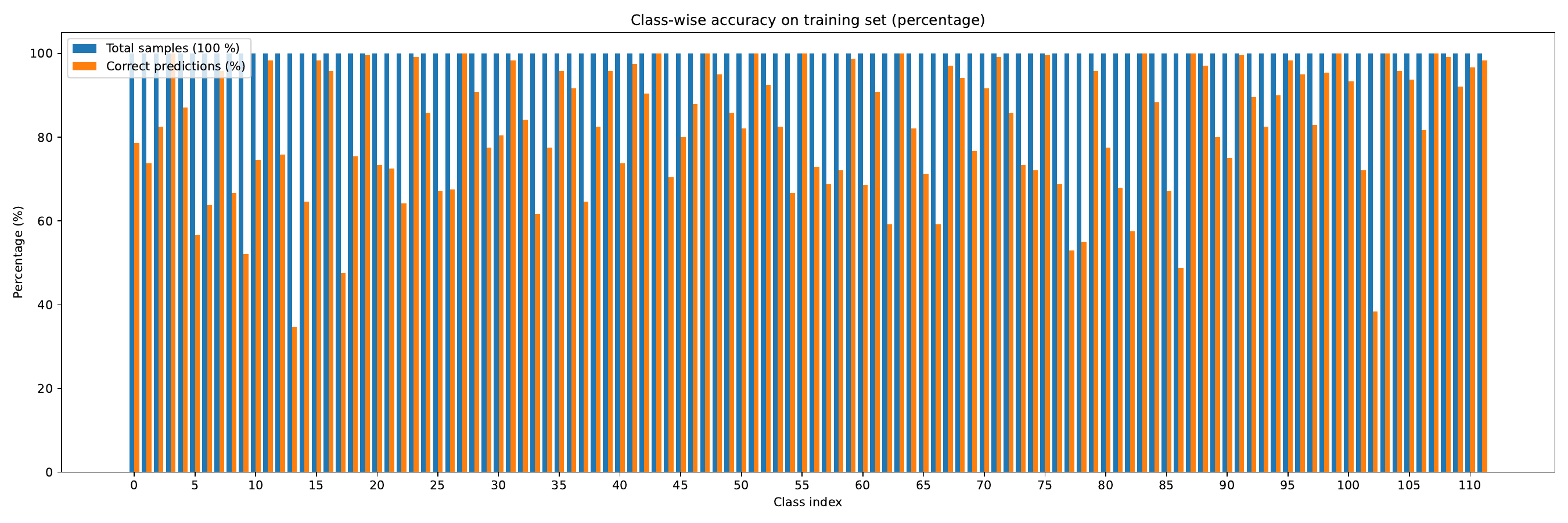}
    \caption{Percentage of correct predictions on train set for all 112 classes of diacritized Arabic letters pronunciation.}
    \label{fig:mlp_prediction_class_test}
\end{figure*}

\begin{figure*}[h]
    \centering
    \includegraphics[width=\textwidth]{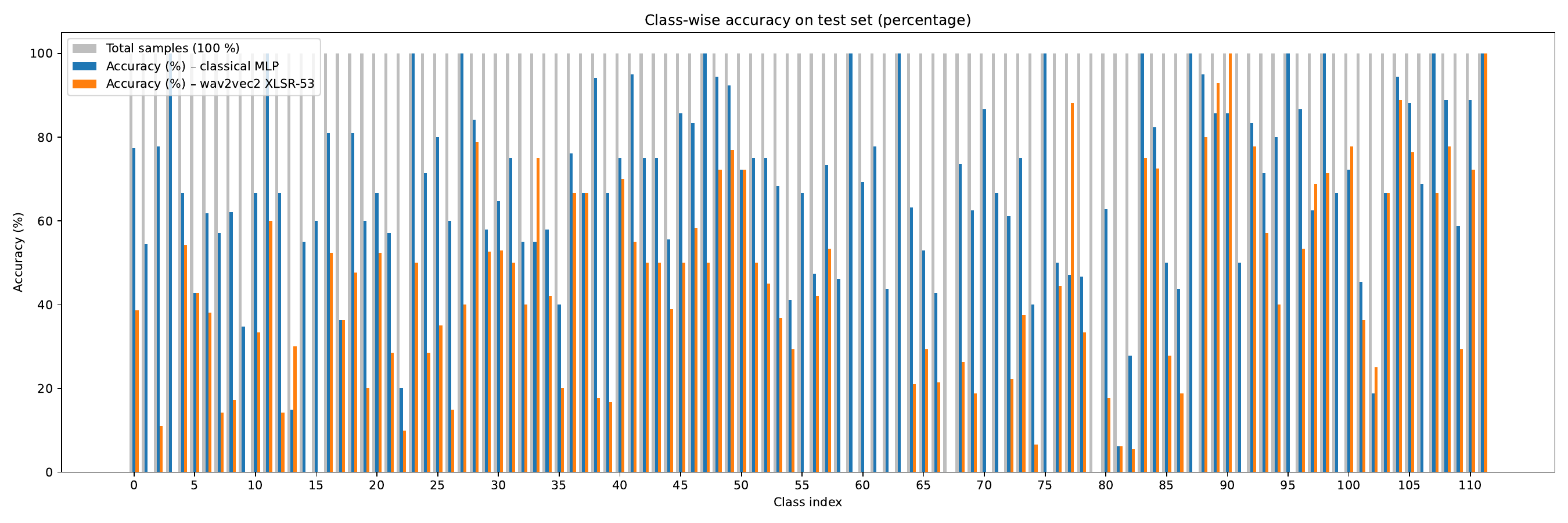}
    \caption{Correct predictions on test set for all 112 classes for a trained classical MLP and state-of-the-art for ``wav2vec2-large-xlsr-53-arabic".}
    \label{fig:mlp_prediction_class_train}
\end{figure*}

To establish a baseline for classification, we trained on ``Horouf" training data, a simple multilayer perceptron (MLP) neural network using PyTorch. We intentionally chose a lightweight, easily-reproducible MLP to serve as a lower-bound reference and to isolate the contribution of the Wav2Vec embedding. The architecture consists of three fully connected layers: the input layer projects to 256 hidden units, followed by a ReLU activation and a dropout layer with a 0.3 rate to reduce overfitting. This is followed by another hidden layer of 128 units, again using ReLU and dropout, and finally a linear output layer that maps to the number of target classes. The model was trained using the Adam optimizer with a learning rate of 1e-3, and cross-entropy loss was used as the training objective. Figure \ref{fig:mlp_classifier} shows an accuracy of 76.6\% on validation and 66\% on test data obtained after 9 epochs. Figure \ref{fig:mlp_prediction_class_test} shows the class-wise accuracy of the MLP classifier on the train set. Figure \ref{fig:mlp_prediction_class_train} shows a class-wise comparison of the prediction accuracy on the test set between the trained MLP classifier and \texttt{wav2vec 2.0 XLSR-53}. The average over classes yields an accuracy of  67.84 \% for the MLP classifier. Although the number of samples is the same for all classes during train and test, notice that there is an imbalance in the prediction accuracy, showing that, based on the ``Horouf" dataset, at phoneme-level, some letters are much harder to predict correctly than others and could require more samples for those specific classes.

The goal of building the classifier was only to serve as a baseline, but not as a practically used classifier model. The obtained accuracy is high, indicating that the data is of good quality. The MNIST \cite{mnist_ref} dataset, considered as a high-quality benchmark in computer vision, with just 10 classes, has a total of 70k samples of digits (0 to 9). The ``Horouf" dataset, in its current form, is therefore considered small to be fit for building a classifier from scratch. Practically speaking, ``Horouf" can help optimise state-of-the-art models to close the gap in Arabic phoneme-level pronunciation detection and classification. For this end, we fine-tuned ``wav2vec2-large-xlsr-53-arabic" using ``Horouf". The fine tuning showed an increase in accuracy on the train data reaching 82 \% on the validation set and 65 \% on the test data.

\section{Stability analysis and robust training}

\textbf{Motivation.} In this section, we analyze the stability and robustness of a classifier trained on the ``Horouf" Arabic letter pronunciation dataset, focusing on its performance under adversarial perturbations. Classically trained neural networks are often vulnerable to small changes in the input data, which can lead to significant drops in accuracy. Our goal is to develop a model that remains resilient to such perturbations, including variations in translation, speed, and noise. To this end, we adopt an adversarial training strategy based on Projected Gradient Descent (PGD) \cite{madry2018towards} a widely used and powerful method for generating adversarial examples in deep neural networks. PGD is applied during training to expose the model to worst-case perturbations within an $\epsilon$-ball, improving its robustness to perturbations. We evaluate the model accuracy before and after PGD-based adversarial perturbations and find that our approach significantly enhances the classifier's stability and resistance to input-level distortions. Robustness is particularly critical in audio‐level letter recognition because real-world pronunciation varies in ways that are hard to capture with clean, studio-quality recordings. Speakers naturally change speaking rate (e.g., rapid conversational speech vs. slow, deliberate articulation), introduce prosodic shifts such as emphasis or lengthening, and exhibit subtle co-articulation effects when a target letter is embedded in different phonetic contexts. In addition, recordings collected “in the wild” are subject to channel distortions (microphone mismatch, room reverberation), environmental noise (traffic, background voices), and even mobile-phone post-processing (automatic gain control, compression). Each of these factors can be modeled as a small but potentially adversarial perturbation in the input waveform that pushes the classifier away from the decision boundary learned during standard training. Training with PGD-generated perturbations that mimic time-stretching, pitch-shifting, and additive noise within an $\epsilon$-ball, we force the network to learn representations that are invariant to realistic variations in speed, pitch, and recording conditions. This yields models that not only resist white-box adversarial perturbations but also maintain consistently high accuracy across speakers, devices, and acoustic environments an outcome that is essential for dependable deployment in language-learning apps, assistive speech tools, and low-resource field recordings.

\textbf{Adversarial training.} Adversarial training \cite{madry2018towards} can be formulated as a min-max optimization problem that aims to improve the robustness of a neural network. Given a model parameterized by $\theta$, a dataset of input-label pairs $(x, y)$, and a perturbation budget $\epsilon$, adversarial training seeks to find model parameters that minimize the worst-case loss within an $\epsilon$-ball around each data point. Formally, the objective is

$$\min_{\theta} \mathbb{E}_{(x, y) \sim \mathcal{D}} \left[  \max_{ ||\delta|| \leq \epsilon}  \mathcal{L}(f_{\theta}(x + \delta), y) \right] $$ 


where $\delta$ is the adversarial perturbation constrained by $||\delta|| \leq \epsilon$, and $\mathcal{L}$ denotes the loss function, such as the cross-entropy loss. The inner maximization problem corresponds to generating adversarial examples that fool the model, while the outer minimization encourages the model to correctly classify these adversarial examples, thereby enhancing its robustness against potential perturbations. The same MLP model is adversarially trained, reaching 67.5\% validation accuracy and 58.96 \% test accuracy. It is well established that adversarially trained models, while more robust to perturbations, typically exhibit lower average test accuracy than the same architectures trained with standard (non-adversarial) procedures.

\textbf{Projected Gradient Descent.} The Projected Gradient Descent (PGD) perturbation is a widely used method for generating adversarial examples that exploit the vulnerabilities of neural networks. PGD is widely acknowledged as the strongest first-order adversarial perturbation. Hence, a model that remains accurate under PGD perturbations is conventionally considered robust to all other gradient-based (and therefore weaker) adversarial perturbations. Starting from an initial perturbation, typically a random point within an $\epsilon$-ball around the input $x$, the PGD perturbation iteratively updates the adversarial example by taking gradient steps to maximize the loss function. Formally, at each iteration $t$, the adversarial perturbation $\delta_t$ is updated as
$$\delta_{t+1} = \Pi_{|\delta| \leq \epsilon} \left( \delta_t + \alpha \cdot \mathrm{sign} \left( \nabla_x \mathcal{L}(f_{\theta}(x + \delta_t), y) \right) \right)$$

where $\alpha$ denotes the step size, $\mathcal{L}$ is the loss function, and $\Pi_{|\delta| \leq \epsilon}$ represents the projection operator that ensures $\delta$ remains within the $\epsilon$-ball. After multiple iterations, PGD generates adversarial examples that are often more effective than single-step perturbations such as FGSM \cite{DBLP:journals/corr/GoodfellowSS14}, thus serving as a powerful tool for evaluating and improving the robustness of machine learning models.

\begin{figure*}[!h]
    \centering
    \includegraphics[scale=0.55]{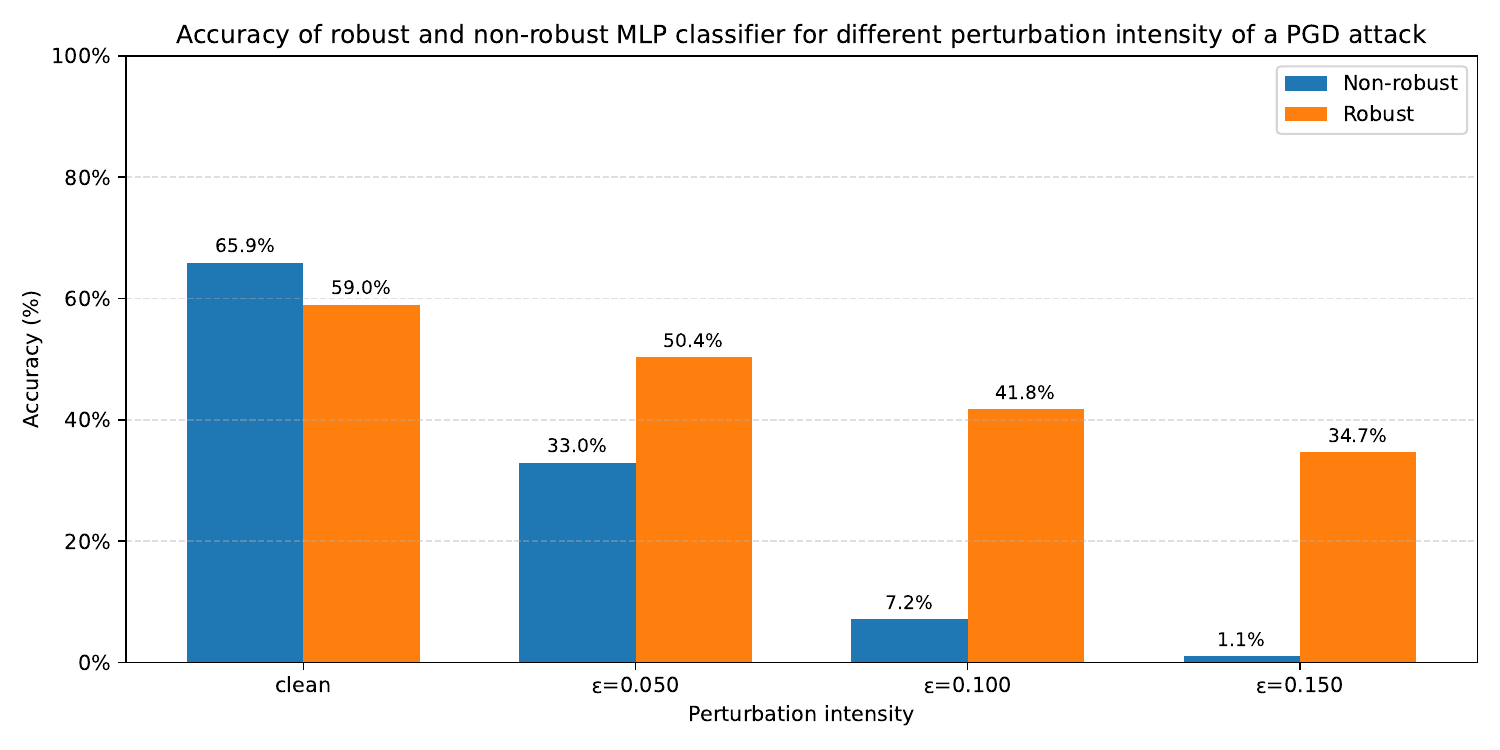}
    \caption{Accuracy comparison between an MLP classifier trained on ``Horouf"  and an adversarially trained classifier having the same architecture, for Projected Gradient Descent (PGD) perturbations with different intensities.}
    \label{fig:adv_training}
\end{figure*}

\textbf{Results.} Figure \ref{fig:adv_training} presents the classification accuracy of the adversarially trained (robust) model versus the vanilla (non-robust) model under Projected Gradient Descent (PGD) attacks of increasing strength. Across all perturbation levels, the robust model shows only a modest decline in accuracy, whereas the non-robust model degrades rapidly as the perturbation amplitude grows. We should observe the same pattern under alternative perturbation schemes, reinforcing the common finding that robustness to PGD tends to generalize to a broad range of perturbations.

\section{Discussion, segmentation and word-level classification}

Current end-to-end transcription models are ill-suited to letter-level pronunciation assessment; they treat words such as ``abi" (``father") and the near-homophone ``api" as indistinguishable because both share a similar acoustic envelope, even though the /b/$\rightarrow$/p/ substitution is a common error among non-native speakers. As a result, these systems fail to deliver per-phoneme feedback and cannot be used to grade articulation quality. To close this gap, we aim to leverage existing Arabic grapheme-level segmenters, well documented in the literature, to partition each diacritized word into its constituent letters (e.g., ``abi" $\rightarrow$ a | bi). We then aim to couple the segmenter with a dedicated letter-based classifier that outputs a confidence score for every segment and an aggregate score for the entire utterance. Extending the segmenter’s pipeline to propagate these letter-specific scores, the combined system would deliver fine-grained feedback that enables language learners,  to pinpoint and remediate individual misarticulations. Our future work will focus on refining this integration across continuous speech, thereby bringing robust letter-level scoring into word- and sentence-level evaluation frameworks without losing the nuanced phonetic distinctions vital to accurate Arabic pronunciation.



\section{Conclusion}

This study supports closing a critical gap in Arabic speech technology: the reliable recognition of isolated letter pronunciations. We showed that state-of-the-art \texttt{wav2vec 2.0} models, while excellent at word- and sentence-level transcription, falter on single-phoneme classification, a weakness that undermines applications in language learning, speech therapy, and phonetic research. To address this, we acquired a carefully curated corpus of isolated letter recordings. The data was collected, via a specially designed mobile and web app, from geographically diverse, gender- and age-balanced  speakers; performed expert-level annotation to establish a high-fidelity ground truth. We fine-tuned \texttt{wav2vec~2.0} on the dataset and a lightweight MLP baseline on this corpus, achieving substantial gains in clean-audio accuracy; and integrated PGD-based adversarial training, boosting robustness against realistic perturbations such as time-stretch, pitch-shift, and environmental noise.

Empirically, we showed that the adversarially trained model maintained high accuracy under $\epsilon$-ball perturbations that cut the baseline’s performance by more than half, and it sharply reduced confusions between phonetically similar letters (e.g., \textsubdot{h}\=a'{} vs h\=a'{}, q\=af vs k\=af). These results confirm that targeted data collection and robustness-oriented training are both necessary and sufficient to deliver dependable letter-level recognition in challenging, real-world conditions.



\paragraph{Acknowledgments}
This study utilized resources from the Core Technology Platforms, the Center for Research Computing at and the Center for Stability, Instability, and Turbulence (SITE) at New York University Abu Dhabi.

\paragraph{Funding Statement}
This research was supported by grants from the Bio-Medical Imaging Core, Core Technology Platforms, New York University Abu Dhabi and the Center for Stability, Instability, and Turbulence (SITE), New York University Abu Dhabi.

\paragraph{Competing Interests}
None

\paragraph{Data Availability Statement}
Dataset and processing pipeline are available upon request. Please submit your request by email to the corresponding author.

\paragraph{Ethical Standards}
The research meets all ethical guidelines, including adherence to the legal requirements of the study country.

\paragraph{Author Contributions} Conceptualization: H.Z; H.H; O.A; N.M. Methodology: H.Z; H.H. Data curation: H.Z. Data visualisation: H.Z. Writing original draft: H.Z. All authors approved the final submitted draft.


\bibliographystyle{unsrt}  
\bibliography{references}

\section*{Authors}
\noindent {\bf Hadi Zaatiti} received the M.E. degree in System Architecture and
Embedded Systems from CentraleSupélec, Paris, France, in 2015, and
the Ph.D. degree in Computer Science from the French Atomic Energy
Commission in 2018. He worked on modeling and formal verification of
hybrid dynamical systems at the French Atomic Energy Commission
from 2015 to 2018. From 2018 to 2022, he was a System Architect and
Data Scientist at the Institute of Research and Technology SystemX. He
then joined Airbus Defence and Space as a System Engineer from 2022
to 2024. Since 2024, he has been a Research Scientist at New York
University Abu Dhabi, leading MEG projects within the Biomedical
Imaging Core.\\

\noindent {\bf Hatem Hajri} received his M.S. and Ph.D. degrees in Applied Mathematics from Paris-Sud University, France, in 2008 and 2011, respectively. From 2011 to 2016, he held teaching and research positions at Paris Nanterre University (Paris 10), the University of Luxembourg, and the University of Bordeaux, where he worked on various topics in stochastic calculus and statistical learning on manifolds. Between 2016 and 2018, he was a researcher at the VeDeCoM Institute in Versailles, focusing on artificial intelligence and autonomous driving. Since 2018, he has been with the Institute for Technological Research SystemX, serving as a lead research scientist on trustworthy AI.\\

\noindent {\bf Osama Abdullah} holds a Ph.D. in Bioengineering from the University of Utah, where his research focused on diffusion tensor MRI of the myocardium. He also holds an M.S. in Bioengineering from the University of Illinois at Chicago and a B.Sc. in Electrical Engineering from the University of Jordan. Since 2017, he has served as an MRI Physicist and Research Instrumentation Scientist at the Bio-Medical Imaging Core of New York University Abu Dhabi, where he applies advanced multi-modal imaging to study brain disorders such as tumors, visual perception dysfunctions, and multiple sclerosis. He also develops imaging analysis tools and safety protocols. Previously, he managed the small animal imaging core at the University of Utah. Dr. Abdullah has received several honors, including the Young Investigator Award from the Department of Education and Knowledge (2020) and merit awards from the ISMRM in 2013 and 2016 for his contributions to MRI research.\\

\noindent {\bf Nader Masmoudi} is a Professor of Mathematics at the Courant Institute of Mathemati- cal Sciences at New York University. His research centers on partial differential equations, nonlinear analysis, and mathematical fluid dynamics, with landmark contributions to the theory of the Navier–Stokes equations, stability of fluid interfaces, and related models in physics. He did  his bachelor at  the Ecole Normale Supérieure in 1996  and received his Ph.D for Paris-Dauphin  in 1999. Professor Masmoudi has held visiting and affiliated positions at leading institutions and has been recognized with several prestigious honors, including election to the American Academy of Arts and Sciences in 2021,  the Kuwait prize 2020, the king Faisal prize 2022, the Fermat prize. \\




\end{document}